\documentclass{article}

\usepackage[final]{corl_2019}

\usepackage[numbers]{natbib}
\usepackage{multicol}
\usepackage{xcolor}
\usepackage{xspace}
\usepackage{graphicx}
\usepackage{booktabs}
\usepackage{pifont} %
\usepackage{balance}
\usepackage{amsmath,amssymb}
\usepackage{breqn}
\usepackage{xcolor}
\usepackage{enumitem}
\usepackage{gensymb}
\usepackage{multirow}
\usepackage{array}
\usepackage{caption}

\usepackage{changepage}

\usepackage{inputenc}
\usepackage[final,tracking=true,kerning=true,spacing=true,stretch=10]{microtype}
\usepackage[USenglish]{babel}
\SetTracking{encoding=*, shape=sc}{0}

\newcommand{\PreserveBackslash}[1]{\let\temp=\\#1\let\\=\temp}
\newcolumntype{C}[1]{>{\PreserveBackslash\centering}p{#1}}
\newcolumntype{R}[1]{>{\PreserveBackslash\raggedleft}p{#1}}
\newcolumntype{L}[1]{>{\PreserveBackslash\raggedright}p{#1}}

\newcommand{\ourmodel}[0]{Discrete Residual Flow Network}
\newcommand{\ourmodelsmall}[0]{discrete residual flow network}
\newcommand{\ourmodelshort}[0]{\textsc{DRF-Net}}
\newcommand{\ourmodelheadshort}[0]{\textsc{DRF}}

\newcommand{\ourmodalitymeasure}[0]{\text{ModePool}}
\newcommand{\modalitythreshold}[0]{\epsilon}

\newcommand{\roadsetshort}[0]{\{\text{CW, ROAD}\}}

\newcommand{\x}[1]{ {\mathbf{x}_{#1}} }
\newcommand{\xbf}[1]{ \mathbf{x} }

\newcommand{\mathbbm}[1]{\text{\usefont{U}{bbm}{m}{n}#1}}
\DeclareCaptionLabelFormat{andtable}{#1~#2  \&  \tablename~\thetable}
\DeclareCaptionLabelFormat{tableandfigure}{\tablename~\thetable  ~\&  \figurename~\thefigure}

\makeatletter
\DeclareRobustCommand\onedot{\futurelet\@let@token\@onedot}
\def\@onedot{\ifx\@let@token.\else.\null\fi\xspace}

\def\eg{\emph{e.g}\onedot} 
\def\ie{\emph{i.e}\onedot}

\makeatother

\hypersetup{draft}

\title{Discrete Residual Flow for Probabilistic\\Pedestrian Behavior Prediction}

\author{\hspace{-1.225cm}Ajay Jain\thanks{Denotes equal contribution.}~ $^{3}$\thanks{Work done while at Uber ATG.}~, Sergio Casas$^{*12}$, Renjie Liao$^{*12}$, Yuwen Xiong$^{*12}$, Song Feng$^{1}$, Sean Segal$^{12}$, Raquel Urtasun$^{12}$\\
\hspace{-1.225cm}Uber Advanced Technologies Group$^1$, University of Toronto$^2$, UC Berkeley$^3$\\
\hspace{-1.225cm}\texttt{ajayj@berkeley.edu}, \texttt{\{sergio.casas,rjliao,yuwen,songf,ssegal,urtasun\}@uber.com}}

\begin{document}
\maketitle

\begin{abstract}
Self-driving vehicles plan around both static and dynamic objects, applying predictive models of behavior to estimate future locations of the objects in the environment. However, future behavior is inherently uncertain, and models of motion that produce deterministic outputs are limited to short timescales. Particularly difficult is the prediction of human behavior. In this work, we propose the \textit{\ourmodelsmall}~(\ourmodelshort), a convolutional neural network for human motion prediction that captures the uncertainty inherent in long-range motion forecasting. In particular, our learned network effectively captures multimodal posteriors over future human motion by predicting and updating a discretized distribution over spatial locations. We compare our model against several strong competitors and show that our model outperforms all baselines.
\end{abstract}


\section{Introduction}

\begin{figure}[t]
    \vspace{-0.7cm}
    \centering
    \includegraphics[width=0.8\linewidth]{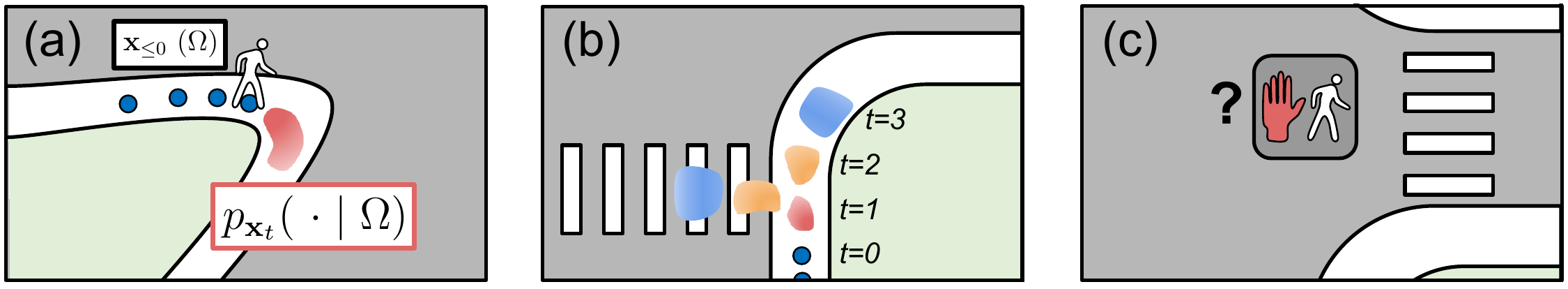}
    \caption{Challenging urban scenarios for pedestrian prediction, depicting pedestrian detections (circles) and future state posteriors colored by time horizon. (a) Gaussian distributions often poorly express scene-sensitive behaviors. (b) Inherent multimodality: the pedestrian may cross a crosswalk or continue along a sidewalk. (c) Partial observability: signals and actors may be occluded.}
    \label{fig:challenges}
\end{figure}

In order to plan a safe maneuver, a self-driving vehicle must predict the future motion of surrounding vehicles and pedestrians. Motion prediction is challenging in realistic city environments. In Figure~\ref{fig:challenges}, we illustrate several challenges for pedestrian prediction. Gaussian distributions often poorly fit state posteriors (Fig.~\ref{fig:challenges}-a). Further, pedestrians have inherently multimodal behavior, as they can move in arbitrary directions and have unknown and changing goals, each achievable with multiple trajectories (Fig.~\ref{fig:challenges}-b). Even with strong evidence for a particular action, such as a road crossing, partially observed environments increase uncertainty in \textit{the timing} of the action (Fig.~\ref{fig:challenges}-c). However, a self-driving vehicle motion planner needs actor predictions to be associated with time. Additional challenges include efficiently integrating spatial and temporal information, the mixed continuous-discrete nature of trajectories and maps, and availability of realistic data.

In the context of self-driving, most prior work represents behaviors through trajectories. Future trajectories can be predicted with a recurrent neural network (RNN) \cite{alahi2016social, becker2018, gupta2018}, a convolutional neural network (CNN) \cite{luo2018fast, casas2018, bansal2018chauffeur}, or with constant velocity, constant acceleration, or expert-designed heuristics. 
However, a trajectory that minimizes the mean-squared error with respect to the true path can only capture the conditional average of the posterior \cite{bishop1994mixture}. The conditional average trajectory does not represent all possible future behaviors and may even be infeasible, lying between feasible trajectories. %

To express multiple possible behaviors, a fixed number of future trajectories can be predicted \cite{cui2018multimodal}, or several can be sampled \cite{gupta2018, sadeghian2019sophie}. Still, in realistic environments, posterior predictive distributions are complex and a large number of samples are needed to capture the space of possibilities. Such models tradeoff prediction completeness and latency from repeated sampling. Further, the number of possible trajectories increases exponentially over long time horizons, and uncertainty grows rapidly.

Instead of predicting trajectories, in this work, we take a probabilistic approach, predicting distributions over pedestrian state at each timestep that can directly be used for cost-based self-driving vehicle planning. Conditioning on a spatio-temporal rasterization of agent histories aligned to the local map, we leverage deep convolutional neural network architectures for implicit multi-agent reasoning, and mimic human dynamics through a \textit{\ourmodelsmall{}}, which we refer to as \ourmodelshort{}. We summarize our contributions as follows:

\begin{itemize}[noitemsep,topsep=0pt]
  \item We develop a deep probabilistic formulation of actor motion prediction that provides marginal distributions over state at each future timestep without expensive marginalization or sampling. Our \textit{discrete residual flow} equation is motivated by autoregressive generative models, and better captures temporal dependencies than time-independent baselines. %
    \item We propose the convolutional \ourmodel{} that predicts actor state over long time horizons with highly expressive discretized distributions.
  \item We thoroughly benchmark model variants and demonstrate the benefit of belief discretization on a large scale, real-world dataset. We evaluate the \textit{likelihood}, \textit{displacement error}, \textit{multimodality}, \textit{entropy}, \textit{semantic mass ratio} and \textit{calibration} of the predictions, using a novel ModePool operator for estimating the number of modes of a discrete distribution. %
\end{itemize}

\vspace{-2mm}
\section{Related work}
\label{sec:related}

Prior work on pedestrian prediction has largely modeled trajectories, goals, or high-level intent.

\textbf{Human trajectory forecasting} ~The pedestrian prediction literature is reviewed in \cite{ridel2018review, Rudenko2019HumanMT}. Multi-pedestrian interactions have been modeled via pooling \cite{alahi2016social, gupta2018} or game theory \cite{ma2017forecasting}. \citet{becker2018} predict future trajectories with a recurrent encoder and MLP decoder, reporting lower error than more elaborate multi-agent schemes, and find that behaviors are multimodal and strongly influenced by the scene. Social GAN \cite{gupta2018} is a sequence-to-sequence generative model where trajectory samples vary in speed and turning angle, trained with a variety loss to encourage diversity. However, the runtime of the sampling approach scales with the number of samples ($150$ ms for $12$ trajectories), even without using a local map, and many samples are needed. SoPhie \cite{sadeghian2019sophie} is another sampling strategy integrating external overhead camera imagery. 
In contrast, we predict entire expressive spatial distributions rather than individual samples and incorporate a local map into prediction.

\textbf{Goal directed prediction}
~\citet{ziebart2009planning} use historical paths to pre-compute a prior distribution over pedestrian goals indoors, then develop an MDP to infer a posterior distribution over future trajectories. \citet{wu2018probabilistic} use a heuristic to identify possible goal locations in a mapped environment and a Markov chain to predict the next-time occupancy grid. %
Rehder et al. \cite{rehder2018pedestrian, rehder2015goal} use a two-stage deep model to predict a Gaussian mixture over goals, then construct distributions at intermediate timesteps with a planning network. 
Still, the number of mixture components must be tuned, and the mixture is discretized  during inference, which is computationally expensive. %
Fisac and Bajcsy \cite{fisac2018, bajcsy2018} specify known goals for each human indoors, then estimate unimodal state distributions by assuming humans approximately maximize utility \ie{} progress toward the goal measured by Euclidean norm. They estimate prediction confidence from model performance and return uninformative distributions at low confidence. Confidence estimation is complementary to our approach.%

\textbf{Semantic map} ~Pedestrian predictors have separately reasoned about spatially continuous trajectories and discretized world representations \cite{ziebart2009planning, rehder2018pedestrian}. These works either ignore the semantic map or integrate it at an intermediate stage. In vehicle prediction, input map rasterizations are more widely used. IntentNet \cite{casas2018} renders a bird's-eye view of the world to predict vehicle trajectories and high-level intention simultaneously, using a rasterized lane graph and a 2D convolutional architecture to improve over previous work \cite{luo2018fast}. Similar map rasterizations are used in \cite{bansal2018chauffeur, yang2018hdnet, djuric2018motion}, and this work. %

\textbf{Related modeling techniques}
~The convolutional long short-term memory (ConvLSTM) architecture has been applied to spatio-temporal weather forecasting \cite{xingjian2015convolutional}. A ConvLSTM iteratively updates a hidden feature map, from which outputs are derived. In contrast, \ourmodelshort{} sequentially adapts the output space rather than a hidden state. Similarly, the adaptive instance normalization operator \cite{huang2017arbitrary} uses a shared feature to predict and apply scale/shift parameters to a fixed, discrete image. Normalizing flows \cite{rezende2015variational} apply a series of invertible mappings to samples from a simple prior, \eg{} a Gaussian, constructing a random variable with a complex PDF. %
While normalizing flows transform individual samples, we directly transform a probability mass function (PMF) for computational efficiency. %

\section{\ourmodel}
\label{sec:drfnet}

In this paper, we express beliefs over future pedestrian positions through categorical distributions that discretize space. Such distributions can be used for cost-based planning or constrained path optimization in self-driving vehicles. In this section, we explain how we represent historical observations as a multi-channel image encoding both the known map and detected actors, a process we call \emph{rasterization}. We then introduce a backbone deep neural network which extracts features from the rasterized image, followed by the probabilistic framework for our \ourmodelshort{}. Finally, we introduce our \ourmodelheadshort{} head which uses the extracted features for prediction.

\begin{figure}[t]
    \vspace{-0.7cm}
    \centering
    \includegraphics[width=\linewidth]{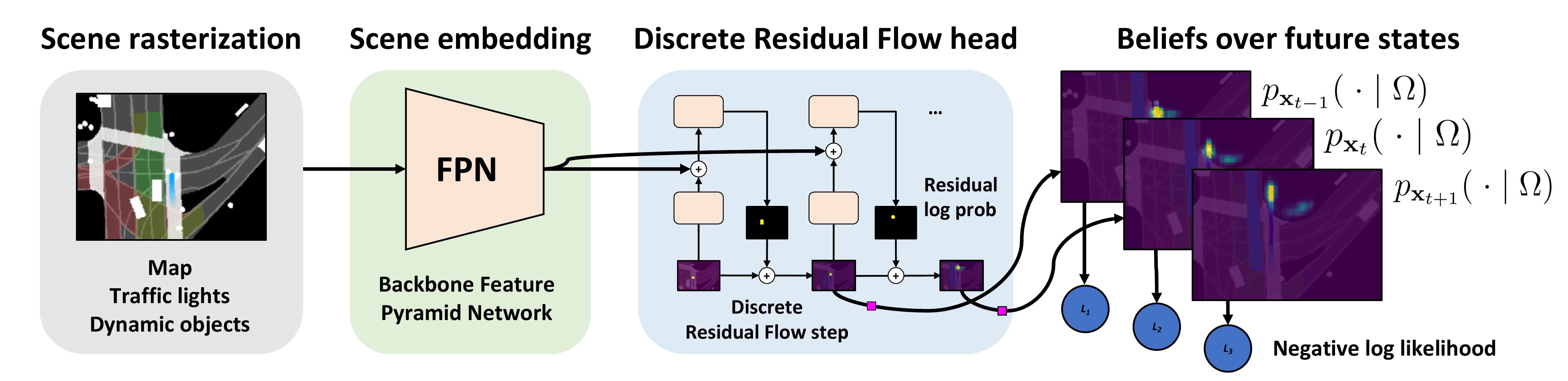}
    \caption{Overview of the \ourmodel{}. Pedestrian of Interest (PoI) and actor detections are aligned with a semantic map. A multi-scale backbone jointly reasons over spatio-temporal information in the input, embedding context into a feature $\mathcal{F}$. Finally, the \ourmodelheadshort{} head recursively adapts an initial distribution to predict future pedestrian states on long time horizons.}
    \label{fig:overview}
\end{figure}

\begin{figure}[t]
    \vspace{-0.7cm}
    \centering
    \vspace{-1mm}
    \includegraphics[width=0.82\textwidth]{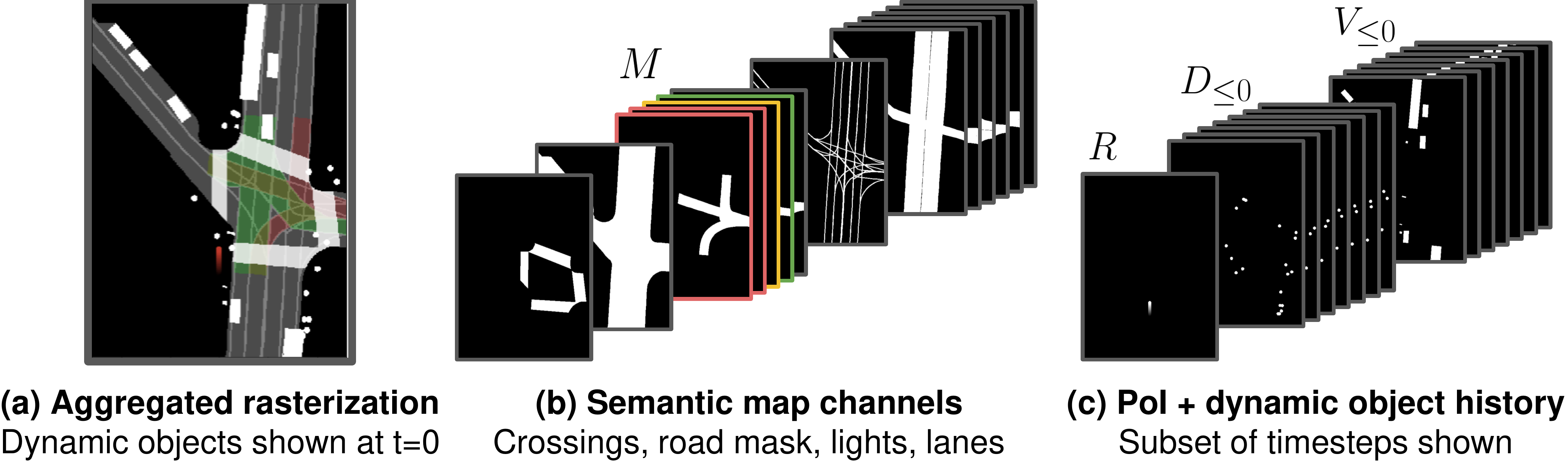}
    \caption{\textbf{Scene history and context representation.} \ourmodelshort{} rasterizes map elements into a shared spatial representation (b), augmented with spatio-temporal encodings of actor motion (c).\vspace{1mm}}
    \label{fig:rasterization}
\end{figure}

\textbf{Encoding Historical Information}\label{sec:rasterization} ~Future pedestrian actions are highly correlated with historical actions. However, actions are also influenced by factors such as road surface types, traffic signals, static objects, vehicles, and other pedestrians. We \textit{rasterize} all semantic map information and agent observations into a 3D tensor, encoding both spatial and temporal information by automatic rendering. The first two dimensions correspond to the spatial domain and the third dimension forms channels. Each channel is an $576 \times 416$ px image encoding specific local bird's eye view (BEV) information at a resolution of $8$ px per meter. Figure~\ref{fig:rasterization} shows an example rasterization from a real urban scene.

Dynamic agents are detected from LiDAR and camera with the object detector proposed in \citet{liang2018deep}, and are associated over time using a matching algorithm. Resulting trajectories are refined using an Unscented Kalman Filter \cite{wan2000unscented}. 
\ourmodelshort{} renders detected pedestrians in each timestep $t$ for the past $6$ seconds in channel $D_t$ and detected non-pedestrians (\eg{} vehicles) in channel $V_t$. To discriminate the pedestrian of interest from other actors, a grayscale image $R$ masks their tracklet.

\ourmodelshort{} renders the local map in a similar fashion to \cite{casas2018}, though centers the map about the PoI. $15$ semantic map channels $M$ finely differentiate urban surface labels. These channels mask crosswalks, drivable surfaces, traffic light state, signage, and detailed lane information. Maps are annotated in a semi-automated fashion in cities where the self-driving vehicle may operate, and only polygons and polylines are stored. 
The final rasterization is $\Omega = \left[ D_{\leq 0}, V_{\leq 0}, R, M \right]$ where $[\cdot]$ indicates concatenation along the channel dimension, the subscript $\leq 0$ indicates a collection of elements from all past timesteps and $t=0$ is the last timestep. All channels are rotated such that the currently observed PoI is oriented toward the top of the scene.

\textbf{Backbone Network}\label{sec:backbone}
~\ourmodelshort{} uses a deep residual network with $18$ convolutional layers (ResNet-18) \cite{he2016deep} to extract features $\mathcal{F}$ from the rasterization $\Omega$. We extract $4$ feature maps at $\frac{1}{4}$, $\frac{1}{8}$, $\frac{1}{16}$ and $\frac{1}{16}$ of the input resolution from ResNet-18. These multi-scale intermediate features are upscaled and aggregated into a $\frac{1}{4}$ resolution global context with a feature pyramid network (FPN) \cite{lin2017feature}.

\textbf{Probabilistic Actor State Prediction} ~We now introduce a probabilistic formulation of future actor state prediction. Given rasterization $\Omega$, we are interested in inferring a predictive posterior distribution over possible spatial locations of the PoI for each timestep $t$ where $t = 1, \cdots, T_{f}$. Instead of treating the state as a continuous random variable, we discretize space to permit a one-hot state encoding. Specifically, we divide space into a grid with $K$ bins. The state at time $t$, $\x{t}$, is a discrete random variable which takes one of the $K$ possible bins.

Consider the joint probability of the states in the future $T_{f}$ timesteps, \ie, $p_{\x1 \cdots, \x{T_{f}}}(x_1, \cdots, x_{T_f} \mid \Omega)$. %
This distribution can be modeled with several factorizations. 
The first and the most straightforward factorization assumes conditional independence of future timesteps,
\begin{equation}
p_{\x1, \cdots, \x{T_{f}}}(x_1, \cdots, x_{T_{f}} \mid \Omega) = {\textstyle\prod}_{t} ~p_{\x{t}}(x_{t} \mid \Omega)
\end{equation} 
We can use a neural network, \eg{}, a CNN, to directly model $p_{\x{t}}(x_{t} \mid \Omega)$. In Section~\ref{sec:experiments_main}, we show the performance of a \textit{mixture density network} and \textit{fully-convolutional predictor} that simultaneously predict these factors. 
Still, conditional independence is a strong assumption. The second factorization follows an autoregressive fashion, providing the foundation for many models in the literature,
\begin{equation}
p_{\x1, \cdots, \x{T_{f}}}(x_1, \cdots, x_{T_f} \mid \Omega) = {\textstyle\prod}_{t} ~p_{\x{t} \mid \x{\leq t-1}}(x_{t} \mid x_{\leq t-1}, \Omega)
\end{equation}
For example, recurrent encoder-decoder architectures \cite{alahi2016social, becker2018, gupta2018} sample trajectories one state at a time and capture the conditional dependencies through a hidden state.

In contrast to the sample-based approach, often we desire access to compact representations of $p_{\x{t}}(x_{t} \mid \Omega)$ for a particular $\Omega$, such as an analytic form or a discrete categorical distribution. As we always condition on $\Omega$, we refer to $p_{\x{t}}(x_{t} \mid \Omega)$ as a marginal distribution. Access to the marginal provides interpretability, parallel sampling and ease of planning as the marginals can be used as occupancy grids. However, direct marginalization is expensive if not intractable as we typically have no simple analytic form of the joint distributions. Approximation is possible with Monte Carlo methods, though many samples are needed to characterize the marginal.

Instead, we propose a \textit{flow} between marginal distributions that resembles an autoregressive model in its iterative nature, but avoids sampling at each step. In contrast to a normalizing flow \cite{rezende2015variational}, which approximates a posterior over a single random variable by iteratively transforming its distribution, discrete residual flow transforms between the marginal distributions of different, temporally correlated random variables by exploiting a shared domain.

\textbf{Discrete Residual Flow}
~Our model recursively constructs $p_{\x{t}}(~\cdot \mid \Omega)$ from $p_{\x{t-1}}(~\cdot \mid \Omega)$,
\begin{equation}
  \log p_{\x{t}}(x_t \mid \Omega) = \log p_{\x{t-1}}(x_{t} \mid \Omega) + \underbrace{\log \psi_{t; \theta_{t}}(x_t, p_{\x{t-1}}(~\cdot \mid \Omega), \Omega)}_{\text{Residual}} - \log Z_t, \label{eq:drf_step}
\end{equation}
where we refer to the second term on the right hand side as the \textit{residual}. $\psi_{t; \theta_{t}}$ is a sub-network with parameter $\theta_{t}$ called the \textit{residual predictor} that takes the marginal distribution $p_{\x{t-1}}(~\cdot \mid \Omega)$ and context $\Omega$ as input, and predicts an elementwise update that is used to construct the subsequent marginal distribution $p_{\x{t}}(~\cdot \mid \Omega)$. 
$Z_t$ is the normalization constant to ensure $p_{\x{t}}(~\cdot \mid \Omega)$ is a valid distribution. Note that the residual itself is not necessarily a valid probability distribution.

Eq.~(\ref{eq:drf_step}) can be viewed as a discrete probability flow which maps from the distribution of $\x{t-1}$ to the one of $\x{t}$. We use deep neural networks to instantiate the probability distributions under this framework and provide a derivation of Eq. (\ref{eq:drf_step}) in the appendix, Section~\ref{sec:supp:derivation}.

For initialization, $p_{\x{0}}(~\cdot \mid \Omega)$ is constructed with high value around our $t=0$ PoI position and near-zero value over other states.
In implementation, the residual predictor is a convolutional architecture that outputs a 2D image, a compact and convenient representation as our states are spatial. This 2D image is queryable at state $x_{t}$ via indexing, as is the updated marginal. Additionally, in implementation, we normalize all marginals at once and apply residuals to the unnormalized potential $\tilde p$,
\begin{equation}
  \log \tilde p_{\x{t}}(x_t \mid \Omega) = \log \tilde p_{\x{t-1}}(x_{t} \mid \Omega) + \log \psi_{t; \theta_{t}}(x_t, \tilde p_{\x{t-1}}(~\cdot \mid \Omega), \Omega) \label{eq:unnorm_drf_step}
\end{equation}

\begin{figure}[t]
    \vspace{-0.7cm}
    \centering
    \includegraphics[width=0.93\linewidth]{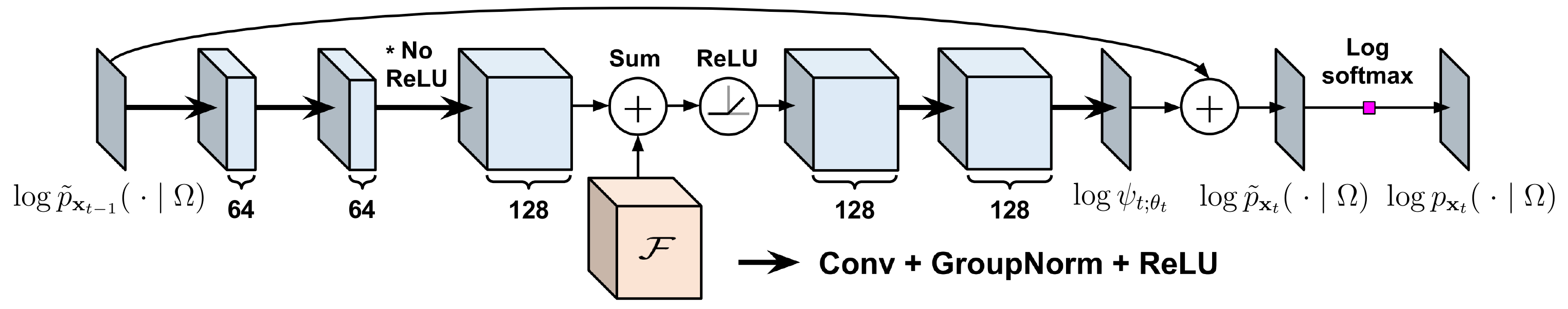}
    \caption{One step of recursive Discrete Residual Flow. The log potential is used to update the global feature map $\mathcal{F}$. \ourmodelheadshort{} then predicts a residual $\psi_{t; \theta_t}$ to flow to the log potential for the next timestep.}
    \label{fig:drf}
\end{figure}

Figure~\ref{fig:overview} illustrates the overall computation process. The embedding of the rasterization $\mathcal{F}(\Omega)$ is shared at all timesteps, used by each residual predictor.
Figure \ref{fig:drf} further illustrates the architectural details of the \ourmodelheadshort{} residual predictor for one timestep.

\textbf{Learning} ~We perform learning by minimizing the negative log likelihood (NLL) of the observed sequences of pedestrian movement. Specifically, we solve the following optimization,
\begin{equation}
    \min_{\Theta} \qquad - \mathbb{E}_{\textbf{x}, \Omega} \left[ {\textstyle\sum}_{t=1}^{T_{f}} \log p_{\x{t}}(x_{t} \mid \Omega) \right]
    \label{eq:nll}
\end{equation}
where the expectation $\mathbb{E}\left[ \cdot \right]$ is taken over all possible sequences and will be approximated via mini-batches.
$\Theta = \{ \theta_{1}, \cdots, \theta_{T_{f}}, w \}$ where $w$ denotes the parameters of the backbone network.

\section{Evaluation}
\label{sec:experiments}

There is not a standard dataset for probabilistic pedestrian prediction with real-world maps and dynamic objects. Thus, we construct a large-scale dataset of real world recordings, object annotations, and online detection-based tracks. We implement baseline pedestrian prediction networks inspired by prior literature \cite{saleh2018long, zyner2018naturalistic, rehder2018pedestrian} and compare \ourmodelshort{} against these baselines on standard negative log likelihood and displacement error measures. We propose an evaluation metric for measuring prediction multimodality, which is one of the most characteristic properties of pedestrian behavior. We also analyze the calibration, entropy and semantic interpretation of predictions. Finally, we present qualitative results in complex urban scenarios.

\subsection{Dataset} \label{sec:dataset}
Our dataset consists of 481,927 ground truth pedestrian trajectories gathered in several North-American cities. The dataset is split into 375,700 trajectories for training, 34,571 for validation, and 71,656 held-out trajectories for testing. Dynamic objects are manually annotated in a $360\degree$, $120$~m range view from an on-vehicle LiDAR sensor. Annotations contain $6$~s ($30$ frames) of past observations and $10$~s ($50$ frames) of the future. These $5$ Hz, $16$~s sliding windows are extracted from longer logs.

We also fine-tune and evaluate \ourmodelshort{} with variable length trajectories from an object detector in the same scenarios. The detector is discussed in Section~\ref{sec:rasterization}. This assesses real-world, on-vehicle prediction performance, reflecting the challenges inherent to real perception such as partial observability, occlusion and identity switches in tracking algorithms. %
While PoIs are annotated for a full $16$ seconds in our ground truth experiments, realistic tracks are of variable length. A self-driving vehicle must predict the behavior of other agents with a very limited set of observations. Thus, we evaluate \ourmodelshort{} by predicting $10$ seconds ($50$ frames) into the future, given tracks with as few as $3$ historical frames, sufficient for estimating acceleration. Relaxing the requirements about past history avoids skewing our dataset toward easily tracked pedestrians, such as stationary agents.

\subsection{Baselines}
In this section, we describe two baseline predictor families. These baselines are trained end-to-end to predict distributions given features $\mathcal{F}(\Omega)$ produced by the same backbone as our proposed model.

\begin{table}[t]
\centering
\vspace{-1mm}
\setlength{\tabcolsep}{5.7pt}
\small
\begin{tabular}{@{}l|cccc|c|ccc|cc@{}}
\toprule
  & \multicolumn{4}{c|}{Negative log likelihood (NLL)} & ADE (m) & \multicolumn{3}{c|}{FDE (m)} & \multicolumn{2}{c}{Mass Ratio (\%)}\\
 Model            & Mean          & @ 1 s         & @ 3 s         & @ 10 s        & 0.2-10s       & @ 1 s         & @ 3 s         & @ 10 s        & Acc.  & Recall \\ \midrule
 Density Net      & 5.39          & 2.87          & 3.96          & 6.74          & 3.49         & 0.93          & 1.72          & 7.66          & 77.99 & 81.33 \\
 MDN-4            & 3.01          & 1.64          & 2.00          & 4.33          & 1.47         & 0.38          & 0.69          & 3.38          & 87.85 & 84.12 \\
 MDN-8            & 3.43          & 1.60          & 2.77          & 4.79          & 1.78         & 0.60          & 0.88          & 3.91          & 85.56 & 84.19 \\
 ConvLSTM         & 2.51          & 0.89          & 1.86          & 4.07          & 1.58         & 0.47          & 1.06          & 3.20          & 88.02 & 85.02 \\ \midrule
 \ourmodelshort{} & \textbf{2.37} & \textbf{0.76} & \textbf{1.74} & \textbf{3.83} & \textbf{1.23} & \textbf{0.35} & \textbf{0.62} & \textbf{2.71} & \textbf{89.78} & \textbf{85.41}\\ \bottomrule
\end{tabular}
\vspace{1mm}
\caption{Comparison of the baselines and our proposed model \ourmodelshort{} with access to ground-truth observations. Metrics are negative log likelihood in $0.5 \times 0.5$ m$^2$ bin containing future GT position, average displacement error (ADE) and final displacement error (FDE) in meters, and percent of predicted mass. Mean NLL, ADE and the mass ratios are averaged over 50 timesteps, $t=0.2-10$ s.}
\label{tab:metrics_nll}
\end{table}

\begin{figure}[t]
\vspace{-0.7cm}
\begin{minipage}[m]{0.6\linewidth}
\centering
\small
\begin{tabular}{@{}l|cccc@{}}
\toprule
 & \multicolumn{4}{c}{Real detection data (NLL)} \\
 Model              & Mean & @ 1 s & @ 3 s & @ 10 s \\ \midrule
 Density Net        & 5.64 & 1.88 & 4.12 & 7.91 \\
 MDN-4 				& 3.21 & 1.52 & 2.54 & 4.71 \\
 MDN-8 				& 3.21 & 1.53 & 2.55 & 4.73 \\
 ConvLSTM           & 3.14 & 1.54 & 2.51 & 4.64 \\ \midrule
 \ourmodelshort{}  	& \textbf{2.98} & \textbf{1.47} & \textbf{2.39} & \textbf{4.36} \\ \bottomrule
\end{tabular}
\captionof{table}{Probabilistic prediction comparison of the baselines and our proposed model \ourmodelshort{} when noisy detections (online tracks) are observed instead of the ground-truth.}
\label{tab:metrics_nll_detections}
\end{minipage}\hspace{2mm}
\begin{minipage}[m]{0.35\linewidth}
\centering
\includegraphics[width=0.75\linewidth]{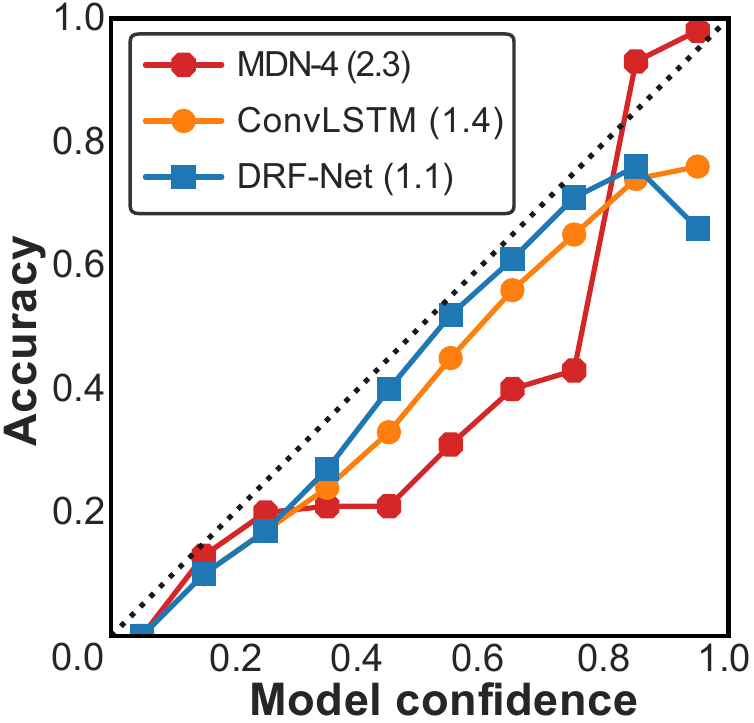}
\caption{Calibration curves and expected calibration error ($*\;10^{-3}\%$)}
\label{fig:calibration}
\end{minipage}
\end{figure}

\textbf{Mixture Density Networks (MDNs)} represent a conditional posterior over continuous targets given continuous inputs with a fully-connected neural network that predicts parameters of Gaussian mixture model \cite{bishop1994mixture}. For a baseline, we implement a variant of this architecture that models pedestrian posteriors at multiple time horizons, conditioned on the past history and current location. Inspired by \citet{rehder2018pedestrian}, we generate the $i$-th mixture component from the neuron outputs $\left\{ m _{ x } , m_ { y } , s _{ x } , s_ { y } , r , p \right\} _ { i }$ which are then reparameterized as $\sigma _{ x , i } = \exp \left( s_ { x , i } \right) + \epsilon, \sigma _{ y , i } = \exp \left( s_ { y , i } \right) + \epsilon , \text { and } \rho _{ i } = \tanh \left( r_ { i } \right)$ to obtain the mean $\vec { \mu } _{ i }$, covariance matrix $\Sigma_ { i }$ and the responsibility of the mixture $\pi_i$: 
\begin{equation}
\vec { \mu } _{ i } = \left[ \begin{array} { c } m_ { x , i } \\ m _{ y , i } \end{array} \right] \text {, } \Sigma_ { i } = \left[ \begin{array} { c c } { \sigma _{ x , i } ^ { 2 } } & { \rho_ { i } \sigma _{ x , i } \sigma_ { y , i } } \\ { \rho _{ i } \sigma_ { x , i } \sigma _{ y , i } } & { \sigma_ { y , i } ^ { 2 } } \end{array} \right] \text {, } \pi _{ i } = \frac{\exp \left( p_ { i } \right)}{ {\textstyle\sum} _{ j = 1 } ^ { N } \exp \left( p_ { j } \right)}
\end{equation}
Training MDNs is challenging due to a high sensitivity to initialization and parameterization. To avoid numerical instabilities, the minimum standard deviation is $\epsilon$. Even with a careful initialization and parameterization, training can be unstable, which we mitigate by discarding abnormally large losses. Note that \citet{rehder2018pedestrian} stabilized training by minimizing only the \textit{minimum} of the batchwise negative log likelihood. Minimizing this minimum loss leads to a good performance on easy examples, but catastrophic performance on hard ones. Lastly, conversions from a discretized spatial input to a continuous output can be challenging to learn \cite{liu2018intriguing}, a problem that our proposed \ourmodelshort{} avoids via a discretized output that is spatially aligned with the input.

\textbf{ConvLSTM}~~ In contrast to our \ourmodelshort{} that recursively updates output distributions in the log-probability space, one can also recurrently update \textit{hidden state} using a Convolutional LSTM \cite{xingjian2015convolutional} that observes the previous prediction. Output distributions are then predicted from the hidden state.

\subsection{Results} \label{sec:experiments_main}
We evaluate negative log likelihood (NLL) at short and long prediction horizons, where lower values indicate more accurate predictions, as well as the mean NLL across all 50 future timesteps. 
In Table~\ref{tab:metrics_nll} and \ref{tab:metrics_nll_detections}, we present results on the held-out test set for ground truth annotated logs and tracked, real-world detections, respectively. Our proposed \ourmodelshort{} achieves a superior likelihood over the baselines by introducing a discrete state representation and a probability flow between timesteps.

\textbf{Likelihood on ground truth tracks} \label{experiments_gt}
~In order to evaluate our results under perfect perception, we benchmark on ground truth (annotated) pedestrian trajectories. Table~\ref{tab:metrics_nll} shows that our proposed model reduces the mean NLL by $0.64$ when compared to the best performer among the MDNs and by $0.14$ with respect to the ConvLSTM baseline. This corresponds to a $90\%$ increase in geometric mean likelihood compared to the best MDN and to a $15\%$ increase when compared to the ConvLSTM.

\textbf{Likelihood on online tracks} \label{experiments_detections}
~Under online, imperfect perception, \ourmodelshort{} achieves a reduction of $0.23$ in mean NLL over the best MDN and $0.16$ over ConvLSTM, \ie{} a $26\%$ and a $17\%$ increase of the geometric mean likelihood of the future observed pedestrian positions, respectively (Table~\ref{tab:metrics_nll_detections}). \ourmodelshort{}'s sequential residual updates may regularize and smooth predictions despite perception noise. Adding more than $4$ components to the density networks does not reduce NLL. Directly predicting occupancy probability over a grid delivers stronger performance than discretizing a continuous spatial density. Using an explicit memory with hidden state updates (ConvLSTM) also has inferior performance to our proposed flow between output distributions.

\begin{figure}[t]
\vspace{-0.7cm}
    \centering
    \vspace{-1mm}
    \includegraphics[width=\textwidth]{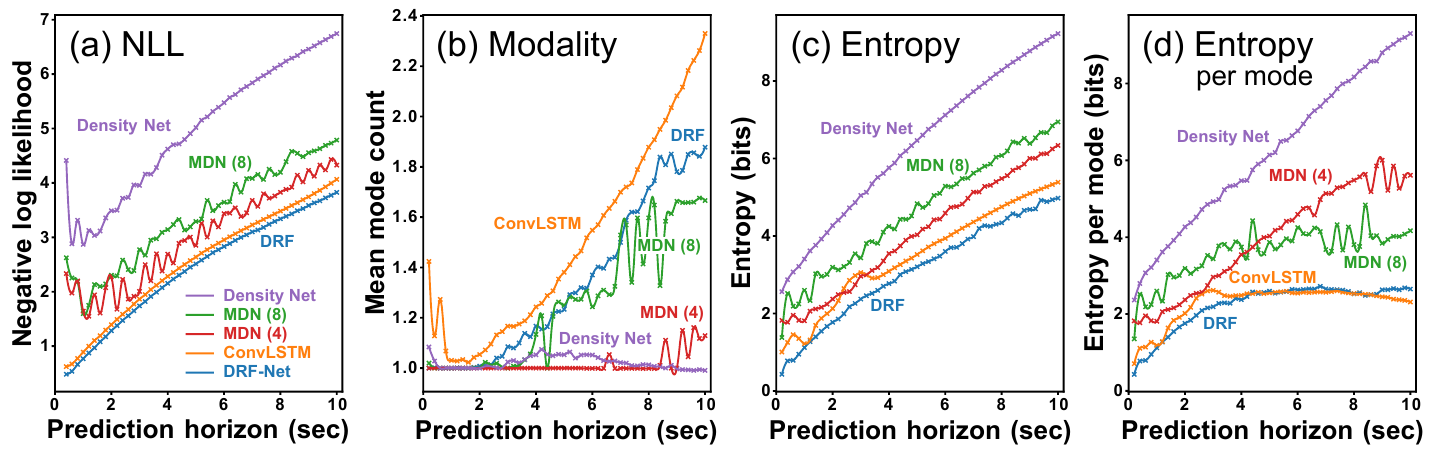}
    \caption{Test metrics. \ourmodelshort{} has low NLL (a) and captures the multimodality inherent in long-range futures (b). Discrete state space (\ourmodelheadshort{}, ConvLSTM) yields the lowest NLL and entropy (c), and entropy per mode saturates. However, EPM increases with horizon for continuous MDNs (d).}
    \label{fig:entropy_modality}
\end{figure}

\begin{table}[t]
\vspace{-1mm}
\centering
\setlength{\tabcolsep}{5.5pt}
\small
\begin{tabular}{@{}l|cccc|cccc@{}}
\toprule
  & \multicolumn{4}{c|}{Ground truth data (NLL)} & \multicolumn{4}{c}{Real detection data (NLL)} \\ 
  Ablative model variant & Mean & @ 1 s & @ 3 s & @ 10 s & Mean  & @ 1 s & @ 3 s & @ 10 s \\ \midrule
 Independent, categorical (Fully conv)		& 2.45 & 0.80 & 1.83 & 3.89 & 3.06 & 1.49 &  2.46	& 4.45\\
 ~+ Sequential refinement (DRR) & 2.40 & 0.80 & 1.78 & \textbf{3.83} & 3.02 	& 1.49 & 2.44& 4.40\\
 Discrete Residual Flow  		& \textbf{2.37} & \textbf{0.76} & \textbf{1.74} & \textbf{3.83} & \textbf{2.98} & \textbf{1.47} & \textbf{2.39} & \textbf{4.36}\\ \bottomrule
\end{tabular}
\vspace{1mm}
\caption{Ablation study of multiple probabilistic prediction heads.  Metric is NLL as in Table \ref{tab:metrics_nll}.}
\label{tab:ablation}
\end{table}

\textbf{Displacement error} ~We compute the expected root mean squared error, or expected displacement error, between the ground truth pedestrian position and model predictions. This is approximated by discretizing posteriors, computing the distance from each cell to the ground truth, and taking the average weighted by confidence at each cell. Table~\ref{tab:metrics_nll} reports the error in meters, averaged over 50 timesteps (ADE) and at specific horizons (FDE). \ourmodelshort{} significantly outperforms all baselines.

\textbf{Model calibration} ~To understand overconfidence of predictive models, we compute calibration curves and expected calibration error (ECE) on the ground truth test set according to \citet{Guo:2017:CMN:3305381.3305518} by treating models as multi-way classifiers over space. ECE measures miscalibration by approximating the expected difference between model confidence and accuracy. DRF has the lowest calibration error, with accuracy closest to the model confidence on average, as shown in Fig.~\ref{fig:calibration}. While somewhat overconfident, these models could be recalibrated with isotonic regression or temperature scaling.

\textbf{Multimodality and Entropy Analysis} ~We propose a \ourmodalitymeasure{} operator to estimate the number of modes of a discrete spatial distribution. \ourmodalitymeasure{} approximates the number of local maxima in a discrete distribution $p$ as follows, where the max is taken over $\left|\delta_r\right|, \left|\delta_c\right| \leq \lfloor \frac{k}{2} \rfloor$, \ie{} $k \times k$ windows:
\begin{equation}
\ourmodalitymeasure{}_{k, \epsilon}(p) = {\textstyle\sum}_{i,j} \mathbbm{1}_{p_{i,j} = \max p_{i+\delta_r, j+\delta_c}} \mathbbm{1}_{p_{i,j} \geq \modalitythreshold{}}
\end{equation}
Only local maxima with mass exceeding a threshold $\epsilon$ are counted. \ourmodalitymeasure{} is efficiently implemented on GPU by adapting the MaxPool filter commonly used in CNNs for downsampling. 
In Figure~\ref{fig:entropy_modality}-b, modality is estimated with $k=5$, $\epsilon=0.1$. Given our output resolution, at most one mode per $2.5 \times 2.5$ m$^2$ area can be counted. While the baseline MDN-4 predicts multiple Gaussian distributions, we observe strong mode-collapse. In contrast, \ourmodelheadshort{} produces predictive posteriors that have increasingly multimodal predictions over horizons. Though an MDN of 8 mixtures captures some multimodality as well, the mean number of modes is highly inconsistent over time (\ref{fig:entropy_modality}-b, middle).

\begin{figure*}[t]
\vspace{-0.7cm}
\centering
\includegraphics[width=\linewidth]{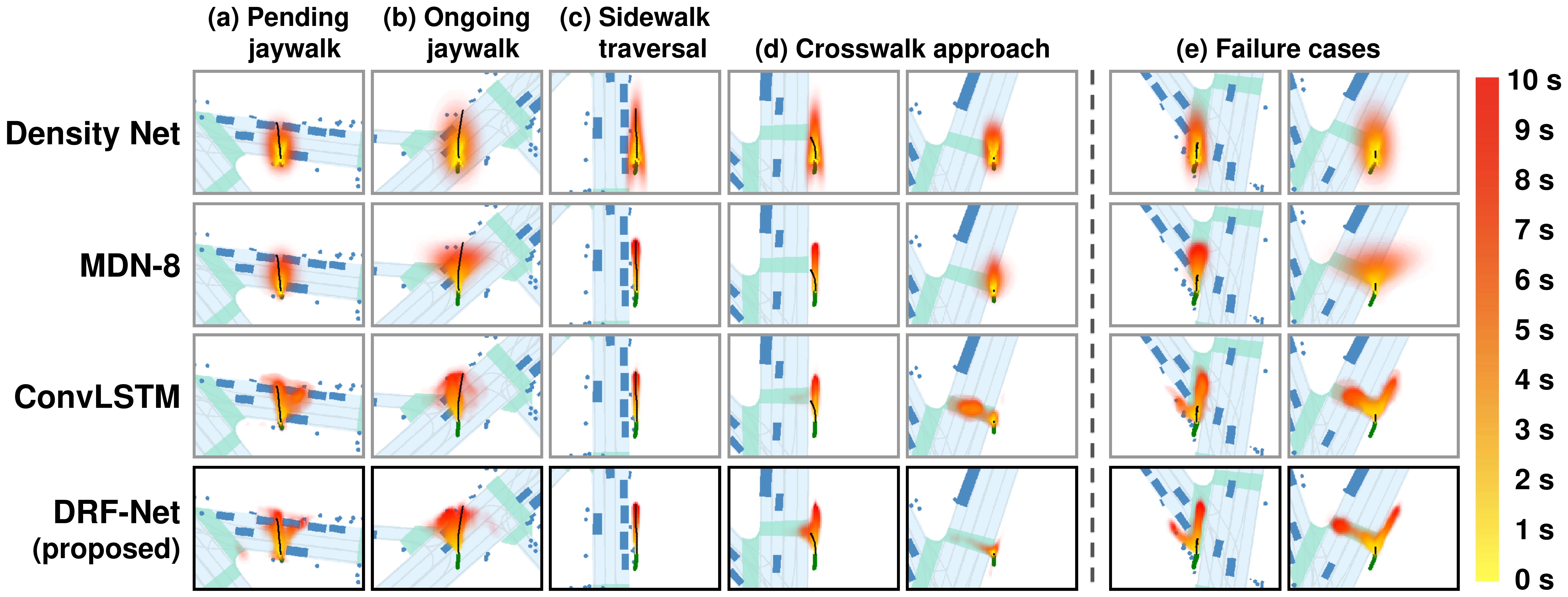}
    \caption{Pedestrian predictions: ground truth past trajectory is green, future is black, opacity shows density, and color shows time horizon. MDN-4 predictions are omitted due to similarity to MDN-8; both are largely unimodal. More results in the supplementary video.}
\label{fig:qualitative}
\end{figure*}

Fig.~\ref{fig:entropy_modality}-c shows the mean entropy of the predicted distributions. Entropy for \ourmodelshort{} is the lowest. As \ourmodelshort{} also achieves lower NLL at all future horizons (\ref{fig:entropy_modality}-a), \ourmodelshort{} predictions can be interpreted as low bias and low variance. We combine entropy and modality into a single metric in Fig. \ref{fig:entropy_modality}-d. For the discrete heads (\ourmodelheadshort{}, ConvLSTM), the entropy per mode saturates. These models capture inherent future uncertainty by adding distributional modes \textit{e.g.} high level actions rather than increasing per-mode entropy. This is not the case for baselines, where entropy per mode grows over time. Qualitatively, in Fig.~\ref{fig:qualitative}, \ourmodelshort{} predictions remain the most concentrated over long horizons.

\textbf{Semantic mass ratio} ~\label{experiments_density_ratio}
Our semantic map can partition the world into three disjoint high-level classes, $\mathcal{C} = \{\text{Crosswalk}, \text{Road}, \text{Off-Road}\}$. To interpret how well models understand the map, we measure \textit{confidence-weighted semantic accuracy}, the mean predicted mass that falls on the correct map class. We also measure \textit{safety-sensitive recall}, the mean mass that falls into a drivable region when the PoI is in a drivable region---performance when a PoI is on-road is very important to a self-driving car. Let $c(\textbf{x}) \in \mathcal{C}$ be the class of location $\textbf{x}$, determined by the map, and $c^{*}_t$ be the ground truth class of the PoI position at time $t$. %
Then, we compute metrics as follows, reported in Table~\ref{tab:metrics_nll}: 
\begingroup
\allowdisplaybreaks
\begin{align}
\text{Accuracy}(c_{\geq 1}^*, \textbf{x}_{\leq 0}, \Omega) &= \frac{1}{T_f} {\textstyle\sum}_{t=1}^{T_f} P(c(\textbf{x}_t) = c_t^*)\\
\text{Recall}(c_{\geq 1}^*, \textbf{x}_{\leq 0}, \Omega) &= \frac{1}{|\text{SS}|} {\textstyle\sum}_{t \in \text{SS}} P(c(\textbf{x}_t) \in \roadsetshort{}),
\end{align}
\endgroup
where $\text{SS} = \left\{ t : c_t^* \in \roadsetshort{} \right\}$, the safety-sensitive timesteps. \ourmodelshort{} significantly outperforms baselines on semantic mass ratio metrics, and most accurately predicts the type of surface the PoI will traverse. This suggests that \ourmodelshort{} better uses the map, and is qualitatively reflected by low-entropy, concentrated mass within map polygons in Figure~\ref{fig:qualitative}.

\textbf{Ablation Study} ~We conduct an ablation study that evaluates the value of discrete predictions and our residual flow formulation. We study two variants of the \ourmodelheadshort{} prediction head, a fully convolutional and a discrete residual refinement (DRR) head. 
MDNs predict continuous mixtures of Gaussians assuming conditional independence of future states, which can be discretized for cost-based planning. We can instead directly predict independent discrete distributions. The fully convolutional predictor projects the spatial feature $\mathcal{F}$ (Section~\ref{sec:backbone}) into a 50-channel space representing per-timestep logits with a 1x1 convolution on scene features. Spatial softmax produces valid distributions over the discrete spatial support. The DRR head takes as input the discrete probability distributions output by our fully convolutional predictor and sequentially predicts per-timestep residuals in log-space with per-timestep weights. DRR thereby refines independent predictions sequentially.

Table \ref{tab:ablation} shows that state space discretization and categorical prediction (fully convolutional head) has significantly better NLL than the best continuous mixture model in Table~\ref{tab:metrics_nll}, a $0.56$ reduction in NLL. Sequential refinement of independent predictions using DRR improves performance. However, predicting flow in the log probability space with DRF achieves the best likelihood.

\textbf{Qualitative Results} ~Figure~\ref{fig:qualitative}-a shows predictions for a pedestrian in a challenging pre-crossing scenario. Predictive posteriors modeled by \ourmodelshort{} (4th row) express high multimodality and concentrated mass, with three visible high-level actions: stopping, crossing straight, or crossing while skirting around a car. \ourmodelshort{} also exhibits strong map interactions, avoiding parked vehicles. However, MDNs predict highly entropic, unimodal distributions, and the ConvLSTM places substantial spurious mass on parked vehicles. Across other test scenes, we observe that \ourmodelshort{} constructs low-entropy yet multimodal predictions with similarly strong map and actor interactions. In Figure~\ref{fig:qualitative}-d, \ourmodelshort{} is the only model to correctly predict a crosswalk approach. Still, in failure cases, all models predict crossings too early, possibly due to unknown traffic light state. This could lead to more conservative self-driving vehicle plans if the pedestrians were nearby. Nonetheless, these pedestrians and lights are distant.

\vspace{-0.3cm}
\section{Conclusion} 
\label{sec:conclusion}
\vspace{-0.3cm}
In this paper, we develop a probabilistic modeling technique applied to pedestrian behavior prediction, called Discrete Residual Flow. We encode multi-actor behaviors into a bird's eye view rasterization aligned with a detailed semantic map. Based on deep convolutional neural networks, a probabilistic model is designed to sequentially update marginal distributions over future actor states from the rasterization. We empirically verify the effectiveness of our model on a large scale, real-world urban dataset. Extensive experiments show that our model outperforms several strong baselines, expressing high likelihoods, low error, low entropy and high multimodality. The strong performance of \ourmodelshort{}'s discrete predictions is very promising for cost-based and constrained robotic planning.

\vspace{-0.3cm}
\section*{Acknowledgments}
\vspace{-0.3cm}
We would like to thank Abbas Sadat for useful discussions during the development of this research.

\bibliography{references}  %

\clearpage
\resetlinenumber

\section{Appendix}

In this appendix, we provide additional implementation (Section \ref{sec:supp:backbone}-\ref{sec:supp:rasterization}), training (Section \ref{sec:supp:training}) and evaluation (Section \ref{sec:supp:metrics}) details for our proposed \ourmodelshort{} and baseline architectures. We also provide a derivation of the DRF update equation (Section \ref{sec:supp:derivation}).

\subsection{Backbone network}
\label{sec:supp:backbone}

\begin{figure}[h]
    \centering
    \includegraphics[width=\linewidth]{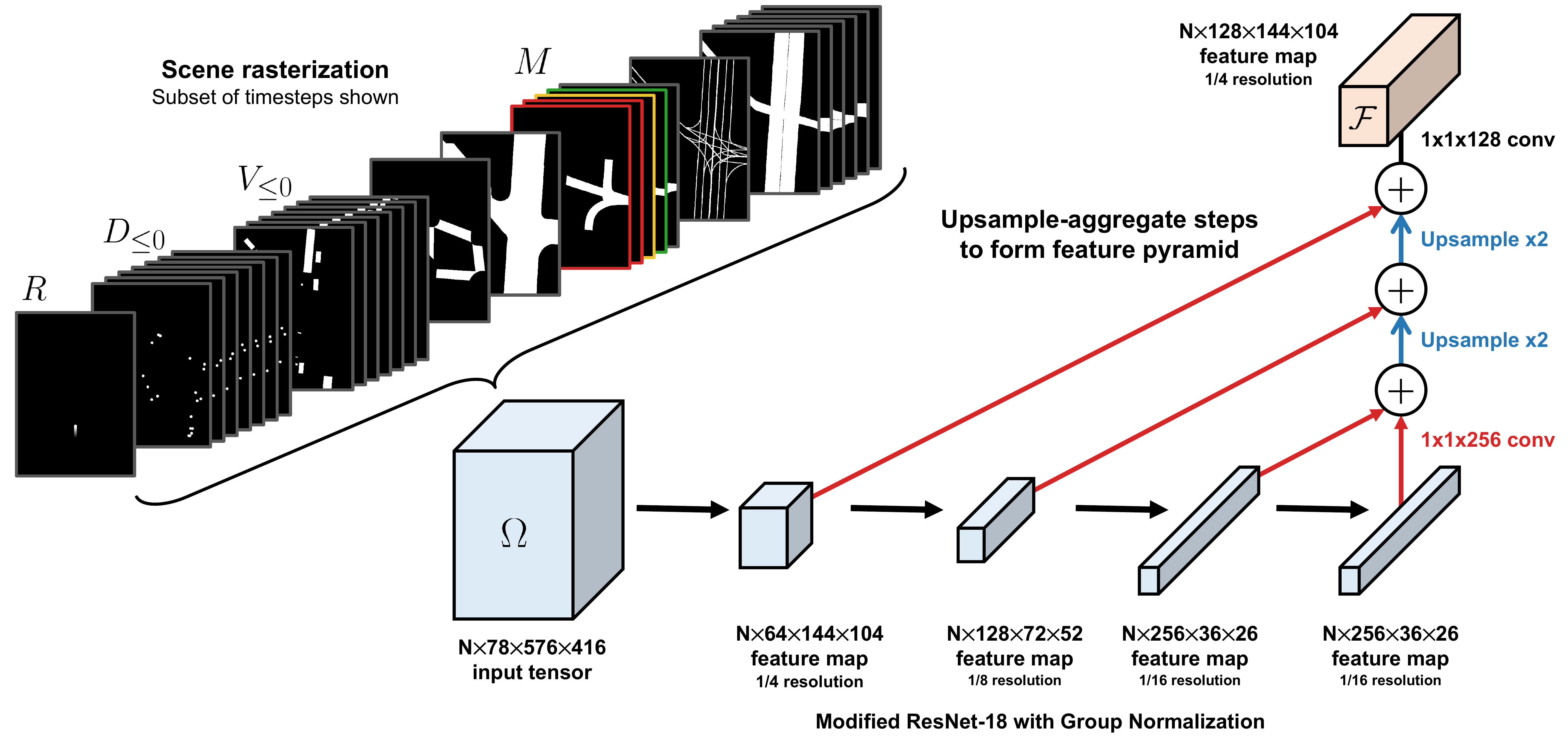}
    \caption{Backbone feature pyramid network (FPN). $N$ denotes the batch size, \eg{} the number of pedestrians of interest for inference or number of scenarios per batch for training.}
    \label{fig:backbone}
\end{figure}

In Section~3 of the paper, we described a deep convolutional neural network architecture that represents our spatio-temporal scene rasterization $\Omega$ as a global feature $\mathcal{F}$. This CNN architecture forms the initial layers of the proposed model and baselines, though each network is trained end-to-end (backbone parameters are not shared across models). The backbone architecture is detailed in Figure~\ref{fig:backbone}, below. The proposed \ourmodelshort{} further projects the $N\times256\times144\times104$ feature $\mathcal{F}$ into a $128$ channel space with a learned $1\times 1$ convolutional filter for memory efficiency.

\subsection{Rasterization}
\label{sec:supp:rasterization}

\paragraph{Rasterization dimensions} The input bird's eye view (BEV) region is rotated for a fixed pedestrian of interest heading at the current time and spans $52$ meters perpendicularly and $72$ meters longitudinally, $50$ ahead and $22$ behind the last observed pose of the pedestrian. We set the input resolution to $0.125$ meters per pixel and the output resolution of our spatial distribution to $0.5$ meters per pixel. At the input resolution, our BEV rasterization channels are each $576$ px by $416$ px.

\textbf{Encoding observed actor behavior}
~We use the object detector proposed in \citet{liang2018deep}, which exploits LiDAR point clouds as well as cameras in an end-to-end fashion in order to obtain reliable bounding boxes of dynamic agents. Further, we associate the object detections using a matching algorithm and refine the trajectories using an Unscented Kalman Filter \cite{wan2000unscented}. 
These detections are rasterized for $T_p=30$ past timesteps, with $200$ ms elapsing between timesteps. At any past time $t$, \ourmodelshort{} renders a binary image $D_t$ for pedestrian occupancy where pixel $D_{t, i, j}=1$ when pixel $i, j$ lies within a convex, bounding octagon of a pedestrian's centroid. Other cells are encoded as $0$. Bounding polygons of vehicles, bicycles, buses and other non-pedestrian actors are also rendered in a binary image $V_t$. In Figure~3-c and Figure~\ref{fig:backbone}, we show how temporal information is encoded in the channel dimension of tensors $D$ and $V$.

To discriminate the pedestrian of interest (PoI) from other actors, a grayscale image $R$ masks the tracklet of the pedestrian to be predicted. As a convention, let the current timestep be $t=0$. If a pixel $i, j$ is contained within the bounding polygon of the PoI at timestep $t \leq 0$, then $R_{i, j} = 1 + \gamma t$, $\gamma \in (0, T_p^{-1})$. By doing so, the whole PoI tracklet is encoded in a single channel with decaying intensity for older detections. This encoding allows for variable track lengths. All rasterization channels are rotated for fixed PoI orientation at $t=0$. We compute orientation with the difference of the last two observed locations.

To allow the network to localize objects in the rasterization, two additional positional encoding channels encode $x$ and $y$ coordinates as real values from $-1$ to $1$, with value $0$ at the last known PoI location. Similar channels are used in \cite{liu2018intriguing}.

\textbf{Encoding semantic map}
~To represent the scene context of the pedestrian, \ourmodelshort{} renders map polygons into $15$ semantic map channels, collectively denoted as $M$, where each channel corresponds to a finely differentiated urban surface label. Crosswalks and drivable surfaces (roadways and intersections) are rasterized into separate channels.
While sidewalks are not explicitly encoded, non-drivable surfaces are implied by the road map. Three channels indicate traffic light state, classified from the on-vehicle camera with a known traffic light position: the green, red, and yellow light channels each fill the lanes passing through intersections controlled by the corresponding light state. Similarly, lanes leading to yield and stop signs are encoded into channels. Finally, we encode other detailed lanes, such as turn, bike, and bus lanes, and a combined channel for all lane markers. In detail, the 15 channels are as follows:%
\begin{enumerate}
    \item Aggregated road mask, masking all drivable surfaces
    \item Masked crosswalks
    \item Masked intersections
    \item Masked bus lanes
    \item Masked bike lanes
    \item All lane markers / dividers
    \item Masked lanes leading to stop sign
    \item Masked lanes leading to yield sign
    \item Lanes controlled by red stop light
    \item Lanes controlled by yellow light
    \item Lanes controlled by green light
    \item Lanes without a turn
    \item Right-turn lanes
    \item Protected left-turn lanes
    \item Unprotected left-turn lanes
\end{enumerate}
This information is annotated in a semi-automated fashion in cities where the self-driving vehicle may operate (Section~\ref{sec:dataset}), and only polygons and polylines are stored.

\subsection{Training}
\label{sec:supp:training}
 
\textbf{Computing negative log likelihood} ~For density visualization in Figure~\ref{fig:qualitative}, and for computing discrete negative log likelihood metrics in Table~\ref{tab:metrics_nll}, the MDN predicted mixture is numerically integrated by a centered approximation with $9$ sampling points for each output grid cell of size $0.5 \times 0.5$ squared meters. Discretizing the MDN allows an NLL metric to be compared between continuous predictions and discrete predictions.

\textbf{Optimization} ~In our experiments with manually annotated trajectories, we train our models from scratch using the Adam optimizer \cite{Kingma2015AdamAM} with a learning rate of $10^{-5}$. When using trajectories from a real perception system, we fine-tune the models learned using the ground truth data to better deal with missing pedestrians and detector/sensor noise. Each training batch includes $2$ pedestrian trajectories. All experiments are performed with distributed training on $16$ GPUs.

\subsection{Metrics}
\label{sec:supp:metrics}

\textbf{Measuring modality} ~To compute the number of modes (local maxima) in a distribution, we proposed the \ourmodalitymeasure{}$_k,\epsilon$ operator. Our proposed operator in fact overestimates modality for MDNs, especially for the Density Network, at short timescales due to quantization error and the fixed window size. To compute modality of a continuous distribution, we discretize the distribution. When the distributions are very long and narrow, as in Density Network short term predictions, multiple modes can be registered. Despite this overestimation, models with the proposed discrete prediction space (ConvLSTM, \ourmodelshort{}) expressed higher multimodality than the MDNs.

\subsection{Derivation of Discrete Residual Flow}
\label{sec:supp:derivation}

We derive Equation (\ref{eq:drf_step}), the discrete residual flow update equation, as an approximation for explicit marginalization of a joint state distribution. %
According to the law of total probability,
\begin{align}
p_{\x{t}} (x_t \mid \Omega) &= \sum_{x_{t-1}} p_{\x{t}, \x{t-1}}(x_{t}, x_{t-1} \mid \Omega) \\
&= \sum_{x_{t-1}} p_{\x{t} \mid \x{t-1}}(x_{t} \mid x_{t-1}, \Omega) ~p_{\x{t-1}}(x_{t-1} \mid \Omega) \label{eq:marginal_update}
\end{align}
Equation~(\ref{eq:marginal_update}) can be seen as a recursive update to the previous timestep's state marginal. Recall that $\x{t}$ is a categorical random variable over $K$ bins. Instead of representing the pairwise conditional distribution $p(x_{t} \mid x_{t-1}, \Omega)$ and conducting the summation once per output bin at $O(K^2)$ cost per timestep, we approximate (\ref{eq:marginal_update}) with a pointwise update,
\begin{align}
p_{\x{t}}(x_t \mid \Omega) &= \sum_{x_{t-1}} p_{\x{t} \mid \x{t-1}}(x_t \mid x_{t-1}, \Omega) ~p_{\x{t-1}}(x_{t-1} | \Omega)\nonumber\tag{\ref{eq:marginal_update}}\\
&= \left[ \sum_{x_{t-1}} \frac{p_{\x{t} \mid \x{t-1}}(x_t \mid x_{t-1}, \Omega) ~p_{\x{t-1}}(x_{t-1} \mid \Omega)}{p_{\x{t-1}}(x_{t} \mid \Omega)} \right] p_{\x{t-1}}(x_{t} \mid \Omega)\\
&\approx \frac{1}{Z_t} \underbrace{\psi_{t; \theta_t}\left(x_t, p_{\x{t-1}}(~\cdot \mid \Omega), \Omega\right)}_{\text{Exponentiated residual}} p_{\x{t-1}}(x_t \mid \Omega) \label{eq:residual_scale}
\end{align}
where $Z_t$ is a normalization constant, and $\psi_{t; \theta_t}$ is a parametric approximator for the summation that we refer to as the \textit{residual predictor}. 
In principle, a sufficiently expressive residual predictor can model the summation exactly. While the residual is applied as a scaling factor in Equation~(\ref{eq:residual_scale}), the residual becomes more natural to understand when the recursive definition is expressed in log domain, completing the derivation,
\begin{equation}
\log p_{\x{t}}(x_t \mid \Omega) = \log p_{\x{t-1}}(x_t \mid \Omega) + \log \psi_{t; \theta_t}\left(x_t, p_{\x{t-1}}(~\cdot \mid \Omega), \Omega\right)  - \log Z_t \label{eq:residual_log}
\end{equation}
We construct $\log \psi_{t;\theta_t}$ such that it can be computed in parallel across all locations $x_t$, and such that the update to $\log p_{\x{t}}(~\cdot \mid \Omega)$ is an elementwise sum followed by normalization. In \ourmodelshort{}, $\log \psi_{t; \theta_t}$ is instantiated with a neural network that outputs a 2D image indexable at these locations (Figure~\ref{fig:drf}). Then, the update (\ref{eq:residual_log}) incurs $O(K)$ cost per timestep.

With this lens, the baseline fully convolutional predictor and the mixture density networks, which assume conditional independence $\x{t}\perp\x{t-1} \mid \Omega$, directly approximate the marginal:
\begin{equation}
\log p_{\x{t}}(x_t \mid \Omega) = \log \psi_{t; \theta_t}\left(x_t, \Omega\right)  - \log Z_t \label{eq:fully_conv_eqn}
\end{equation}
The baseline ConvLSTM propagates a cell and hidden state between steps and shares parameters of the predictor, without sampling from intermediate marginals:
\begin{align}\label{eq:convlstm_update}
\log p_{\x{t}}(x_t \mid \Omega) &= \log f_{\phi}(h_t) - \log Z_t\\
h_t, c_t &= \psi_{\theta}\left(p_{\x{t-1}}(~\cdot \mid \Omega), h_{t-1}, c_{t-1}\right) \nonumber\\
h_0 &= \mathcal{F}(\Omega), c_0\text{ is a parameter} \nonumber
\end{align}
Discrete residual flow retains most of the benefits of the independence assumption, \ie{} tractable marginal distribution estimation and parallelizability, with update more closely resembling a Markov chain. However, there is no sampling between timesteps. As we established in our ablation study (Table~\ref{tab:ablation}), applying DRF Eq. (\ref{eq:residual_log}) outperforms the baseline fully convolutional predictor according to Eq. (\ref{eq:fully_conv_eqn}) and the ConvLSTM update, Eq. (\ref{eq:convlstm_update}).

\end{document}